\documentclass[twoside,11pt]{article}

\usepackage{jmlr2e}
\usepackage{amsmath}
\usepackage{booktabs}
\usepackage[table,xcdraw]{xcolor}
\usepackage[labelformat=simple, caption=false]{subfig} % the labelformat option removes the double parentheses in subfigures
\usepackage[para, online, flushleft]{threeparttable}
\usepackage{tipa}
\usepackage{siunitx}
\usepackage{multirow}
\usepackage{xfrac}
\usepackage{empheq}
\usepackage{svg}

\newcommand{\be}{\begin{equation}}
\newcommand{\ee}{\end{equation}}
\newcommand{\bea}{\begin{eqnarray}} 
\newcommand{\eea}{\end{eqnarray}}

\jmlrheading{1}{2022}{1-18}{1/22}{3/22}{Pablo Ortiz and Simen Burud}

\ShortHeadings{BERT attends the conversation: Improving Low-Resource Conversational ASR}{Ortiz and Burud}
\firstpageno{1}

\begin{document}

\title{BERT Attends the Conversation: \\Improving Low-Resource Conversational ASR}

\author{\name Pablo Ortiz \email pablo.ortiz@telenor.com \\
       \addr Telenor Research\\
       1331 Fornebu, Norway
       \AND
       \name Simen Burud \email simen.burud@gmail.com \\
       \addr Department of Computer Science\\
       Norwegian University of Science and Technology\\
       7491 Trondheim, Norway}

\maketitle

\begin{abstract}%   <- trailing '%' for backward compatibility of .sty file
We propose new, data-efficient training tasks for BERT models that improve performance of automatic speech recognition (ASR) systems on conversational speech. We include past conversational context and fine-tune BERT on trans\-cript disambiguation without external data to rescore ASR candidates. Our results show word error rate recoveries up to 37.2\%. We test our methods in low-resource data domains, both in language (Norwegian), tone (spontaneous, conversational), and topics (parliament proceedings and customer service phone calls). These techniques are applicable to any ASR system and do not require any additional data, provided a pre-trained BERT model. We also show how the performance of our context-augmented rescoring methods strongly depends on the degree of spontaneity and nature of the conversation.
\end{abstract}

\begin{keywords}
  BERT, conversational ASR, rescoring, low-resource ASR, language modelling
\end{keywords}

\section{Introduction}

In recent years, end-to-end Deep Learning-based systems have proved to be very successful, see for example the works of \citet{DBLP:journals/corr/abs-1211-3711,DBLP:journals/corr/ChorowskiBSCB15,amodei2016deep,8068205,8462506,DBLP:journals/corr/abs-1911-08460}. In contrast to HMM-based pipelines, this approach requires very little domain expertise to train since they take the raw audio as input\footnote{Generally, a numerical representation such as MFCCs is extracted from the power spectrum of the Fourier transform of the signal, although there are also attempts to learn input representations using neural networks, see for instance the review of \citet{DBLP:journals/corr/abs-2001-00378}.} and the target transcript as output. All feature extraction and -engineering needed is learned implicitly during training. This makes it relatively easy to adapt an ASR system to a new domain.

Even though these systems are termed end-to-end, most can be decomposed into three distinct components: the acoustic model (AM), the decoder algorithm, and the language model (LM).
The acoustic model is primarily concerned with encoding the audio to a sequence of token probability vectors. This sequence is typically decoded using a wide, compute-intensive beam search, often guided by a language model.
The top-scoring result from the beam search is usually not the optimal one, so the candidate list can be re-scored with a secondary LM with the aim of improving the accuracy of the ranking, see e.g.,  \citep{williams08_interspeech,singh17_interspeech,8461405,DBLP:journals/corr/abs-1910-11450,DBLP:journals/corr/abs-2104-11070,DBLP:journals/corr/abs-2104-04950,DBLP:journals/corr/abs-2106-06922} and references therein.

ASR systems usually do not make use of context beyond the current utterance. This lack of information makes it harder to disambiguate phonetically similar or ambiguous transcripts, and the model often ends up outputting the statistically most common interpretation while disregarding the bigger picture. For this reason there has recently been a significant amount of research to include dialog context in different forms when transcribing a utterance in a dialog both in language modelling \citep{10.5555/1613984.1614018, xiong-etal-2018-session} and ASR \citep{DBLP:journals/corr/abs-1808-02171, DBLP:conf/interspeech/KimDM19, DBLP:journals/corr/abs-2104-11070, Shenoy_2021}. This task is facilitated by the rise of LM pre-training approaches able to compute deeper contextual representations \citep{DBLP:journals/corr/abs-1802-05365, radford2018improving, devlin_bert_2019, brown2020language}.

Even though ASR systems see widespread deployment, they are much less reliable when applied to low-resource languages and real-life situations. One can seek help from general-purpose LMs pre-trained on massive text corpora, which obtain state-of-the-art results in a wide range of NLP tasks with relatively little task-specific fine-tuning \citep{devlin_bert_2019, brown2020language}. These models can be used to aid ASR in a variety of ways, while we focus on using the conversational context learnt by BERT \citep{devlin_bert_2019} for the rescoring of ASR hypotheses \citep{DBLP:journals/corr/abs-1910-11450, DBLP:journals/corr/abs-2104-04950, DBLP:journals/corr/abs-2104-11070, DBLP:journals/corr/abs-2106-06922}. Our work is similar in spirit to that of \citet{DBLP:journals/corr/abs-2104-04950}, however we pursue a different fine-tuning strategy for BERT and apply it to domain-specific, spontaneous conversations in low-resource languages. Our contributions are summarized below.

First, we explore to what extent an LM such as BERT \citep{devlin_bert_2019} improves ASR results after training on a conversational disambiguation task, when the acoustic model (AM) and n-gram LM have only been trained on single, independent utterances, which is our baseline system. We rescore the N-best results with context-augmented BERT, obtaining significant improvements over the baseline. Making conversational context available to BERT in the form of previous utterances is key to our approach. We find that the amount of context required depends on the target domain.

In particular, formal parliamentary discussions greatly benefit from large conversational contexts, while shorter contexts are favoured when rescoring the more unpredictable, spontaneous conversations from our dataset of calls to customer service. Further, we find that fine-tuning procedures play a significant role in BERT's ability to rescore the N-best lists. We propose a data-efficient strategy that uses the baseline ASR system to generate sufficient training examples for BERT, even from a tiny dataset.
Fine-tuning BERT this way on a small number of relevant samples performs far better than doing it on a much larger, out-of-domain conversational dataset.

The outline of the paper is as follows. The next section introduces the baseline ASR system, Section \ref{sec:integration} presents our LM integration approach, Section \ref{sec:training} details our LM training strategies, Section \ref{sec:data} describes the data, Section \ref{sec:results} discusses the experiments and results, and finally in Section \ref{sec:conclusions} we present our conclusions and future work.

%%%%%%%%%%%%%%%%%%%%%%%%%%%%%%%%%%%%%%%%%%%%
%%%%%%%%%%%%%%%%%%%%%%%%%%%%%%%%%%%%%%%%%%%%

\section{Baseline ASR system}
\label{sec:baseline}

Our baseline ASR system is based on Deepspeech 2 \citep{amodei2016deep}, a family of ASR systems based on deep neural networks and the CTC framework \citep{graves_connectionist_2006}. It processes each audio feature only in the context of neighboring features using convolutional layers and then combines them using bidirectional RNN layers. Curriculum learning is applied during the first epoch, using transcript length (in characters) as a proxy for difficulty. We make use of an existing PyTorch implementation\footnote{\url{https://github.com/SeanNaren/deepspeech.pytorch}.} as a code basis for our model.

The model operates at character level. Due to the independence assumption made in CTC, its implicit language model is often found to be weak compared to that of attention-based sequence-to-sequence models, see \citep[e.g.,][]{chan_listen_2016}. Instead, Deepspeech models rely partly on an external n-gram LM during decoding. We train n-gram models using the implementation of Kneser-Ney smoothed n-gram estimation from \citet{heafield_scalable_2013}\footnote{\url{https://github.com/kpu/kenlm} version \texttt{35835f1}.}. We train separate models for each speech corpus and prune unique n-grams of order 2 and above. Each utterance is treated as a separate document, meaning no context information from previous utterances is available in the n-gram models.

We train the \emph{primary} AM with a Norwegian public dataset (NST) consisting of 394.5 hours of recorded and transcribed speech, where 300 hours are used as the training dataset (see Section \ref{sec:data} for further details on the data). The model architecture consists of three strided convolutional layers, nine 1200-dimensional, bidirectional GRU layers, and a fully connected layer with softmax.

As we describe in detail below, to construct the different \emph{baseline} models on other datasets, we fine-tune the primary AM and build specific LMs using the datasets in question.

We test the performance of our primary AM on NST data by combining it with a 5-gram LM in a beam search of size 64.
The LM is trained with approximately 13 million sentences from a non-public corpus gathered by NST consisting of newspaper text.\footnote{We thank August Moum and Skjalg Winnerdal for providing access to this model. It was trained with a version of the newspaper corpus curated at the Norwegian University of Science and Technology (NTNU) during the SVoG project in collaboration with other institutions.}
We achieve a word error rate (WER) of 2.87\% and character error rate (CER) of 1.16\% on the test set, showing that for sufficiently large, clean, generic datasets, these ASR models are rather successful. However, that is certainly not the case when working with small datasets containing noisy, topic-specific, unstructured conversational speech, which motivates the extensions in this work.

In the next section we describe how we augment our baseline system by introducing BERT models. We emphasize that the choice of baseline system is purely circumstantial, and that the extensions discussed in what follows only require the ASR system to produce transcription candidates.

%%%%%%%%%%%%%%%%%%%%%%%%%%%%%%%%%%%%%%%%%%%%
%%%%%%%%%%%%%%%%%%%%%%%%%%%%%%%%%%%%%%%%%%%%
\newpage
\section{N-best rescoring}
\label{sec:integration}

We distinguish between three main elements within the ASR system: acoustic model (AM), beam search (BS) and language model (LM).

The AM takes an audio segment $x$ as input and outputs a matrix $Z$ such that each element $z_k^t$ represents the probability of token $k$ at time $t$. The BS decodes $Z$ into a set $Y$ of the N best candidate transcripts, aided by an n-gram LM through \emph{shallow fusion} \citep{gulcehre_using_2015}. That is,

\begin{equation*}
P_\text{BS}(y | x) = P_\text{AM}(y | x) + \alpha \, P_\text{LM1}(y) + \beta\, \text{WC}(y)\,,
%\label{eq:BS1}
\end{equation*}
%\medskip
\noindent
where $P_\text{AM}$ is the sum of path scores for the transcript as estimated by CTC prefix beam search \citep{graves_connectionist_2006}, $P_\text{LM1}$ is the transcript's probability according to a Kneser-Ney n-gram LM, and WC is a word count bonus.

Rather than training the LM on large external datasets, we obtain
slightly better performance when training it solely on the transcripts used to fine-tune the AM. We thus observe that for domain-specific datasets, relevance (or topic overlap) compensates for abundance\footnote{Our primary AM obtains its best results when combined with an LM trained on abundant external data, but this stems from the fact that the primary AM is trained on data covering an extensive variety of topics, similar to that covered by the external LM.}.  

We denote the highest-ranked transcript from the beam search as
\begin{equation}
y_* \equiv \underset{y \in Y}{\text{argmax}}\ P_{BS}(y | x)\,.
\label{eq:topbeam}
\end{equation}
For the baseline system, $y_*$ is returned as the final transcript, and its WER with respect to the ground truth $y_\text{gt}$ is used as the evaluation metric.

%\bigskip
\noindent
We extend this system to perform N-best\footnote{N $=1024$ throughout the rest of this paper.} rescoring of $y \in Y$ with BERT to obtain the final ranking. To do that, we interpolate the scores from the beam search and the secondary language model (BERT):
\begin{equation}
    P_\text{rescored}(y | x) = (1-\gamma) P_{BS}(y|x) + \gamma\, P_\text{BERT}(y)\,,
    \label{eq:rescoring}
\end{equation}
where $\gamma$ is an adjustable weight parameter. The top candidate after rescoring is denoted by
\be
y' \equiv \underset{y \in Y}{\text{argmax}}\ P_\text{rescored}(y|x)\,.
\ee

N-best rescoring does not make assumptions about the baseline ASR system, making the extension widely applicable. Instead, rescoring makes good use of the many candidate transcripts returned from the wide beam searches typical to CTC-based pipelines.

N-best rescoring exploits the fact that $y_*$ in \eqref{eq:topbeam} is often a suboptimal choice of transcript from $Y$. To identify the upper performance bound, we can define an oracle rescoring algorithm that always selects the best transcript from those present in $Y$:
\begin{equation}
y_o \equiv \underset{y \in Y}{\text{argmin}}\ \text{WED}(y_\text{gt}, y),
\label{eq:y_o}
\end{equation}
where WED is the word-level (Levenshtein) edit distance between two transcripts.

%%%%%%%%%%%%%%%%%%%%%%%%%%%%%%%%%%%%%%%%%%%%
%%%%%%%%%%%%%%%%%%%%%%%%%%%%%%%%%%%%%%%%%%%%

\section{BERT training strategies}
\label{sec:training}

We use BERT as described by \citet{devlin_bert_2019}\footnote{We use the implementation by \citet{Wolf2019HuggingFacesTS}, in particular the model available at \url{https://huggingface.co/NbAiLab/nb-bert-base/tree/6aad23f6c2ed0df8498397175ec07d497f6319a1}, pre-trained in Norwegian.} as our rescoring language model in \eqref{eq:rescoring}.
A challenge in using BERT is that transcripts of spontaneous conversations read very differently from, for example, a novel. People use a very different sentence structure when speaking (such as incomplete sentences and repetitions) and a different choice of words, including fillers and hesitations. In dialogues there are also many cases of overlapping speech and interruptions. Thus the language representations built when training on written text are suboptimal for inference on conversations. To deal with this, BERT needs to be fine-tuned to the task at hand, and also learn spoken language. We propose three different strategies for training BERT, as explained below.

%%%%%%%%%%%%%%%%%%%%%%%%%%%%%%%%%%%%%
\subsection{MLM \& NSP}

The simplest solution is to keep training BERT with the same objective used during pre-training as \citet{devlin_bert_2019}, i.e., Masked Language Modeling (MLM) and Next Sentence Prediction (NSP). These training tasks are known to teach BERT a good language representation. Previous work such as that of \citet{shin_effective_2019} uses a BERT variant trained on these tasks to score sentences and interpolate those LM scores to rescore the hypotheses from a LAS \citep{chan_listen_2016} acoustic model. Using the NSP head for scoring is straightforward, as it already outputs the probability directly. With the MLM output nodes, we can score sentences as follows \citep{shin_effective_2019}:
\begin{enumerate}
    \item Pass the sequence $y=y_1...\,y_L$ through the model, masking the first position.
    \item Among the output probabilities, note the probability of the token at the first position being the original token, denoted $P(y_1 | y_2, ..., y_L)$.
    \item Repeat for each remaining position in the sequence and calculate $P(y) = \prod_{l=1}^L P(y_l | y_{\bar{l}})$, where $y_{\bar{l}}$ denotes all elements in the sequence except $y_l$.
\end{enumerate}

%%%%%%%%%%%%%%%%%%%%%%%%%%%%%%%%%%%
\subsection{Conversational NSP}
\label{sec:cnsp}

In NSP as proposed by \citet{devlin_bert_2019}, BERT's input is fully packed with text, with the separator being placed anywhere inside the input text, dividing it into segments rather than sentences. This reduces BERT's ability to make use of the conversational aspect of our domain.
To remedy this, we adapt the concept of NSP to conversations: positive samples are simply triplets of consecutive utterances, while negative samples are generated by replacing the third utterance with a different one.

\citet{lan_albert_2020} propose a sentence order prediction task with pairs of sentences which creates negative samples by swapping the order of the sentences in the pair. Other objectives are studied by \citet{liu_roberta_2019}, sampling contiguous sentences from the same document or from separate ones. Even though the results of \citet{liu_roberta_2019, lan_albert_2020} improve with respect to the BERT baseline for the datasets they tested, these objectives are not suitable for our spontaneous conversational data: preliminary experiments indicate that sampling from the same conversation is too hard a task. We attribute this to our type of data, where overlapping speech often makes the sentence order ambiguous. In addition, quick turn-taking, short utterances and what appears to be sudden topic changes (without the presence of a much larger context and knowledge base) are ubiquitous. Hence, to make the task more feasible for BERT, we use triplets of utterances and create negative samples by replacing the third utterance by one from another dialogue.

%%%%%%%%%%%%%%%%%%%%%%%%%%%%%%%%%%%%
\subsection{Disambiguation}
\label{sec:disambiguation}

With the training tasks proposed so far, there is an apparent train-inference mismatch: the N-best lists pri\-ma\-ri\-ly contain variations of the same text, often with only a few words differing. Spelling and grammatical errors are also far more common, neither of which are captured by the preceding tasks.

To remedy this, we propose a disambiguation task where BERT's classification head is trained to distinguish the best transcript available from similar candidates. Given a dataset $\mathcal{D}$ of candidate transcripts $Y$ and ground-truth conversational context $c$, we optimize
$$ \min_\theta \sum_{(c, Y) \in \mathcal{D}} \left[ \mathcal{L}\left(1, P_{\text{BERT}_\theta}(\Tilde{y} | c)\right) + \sum_{y \neq \Tilde{y}} \mathcal{L}\left(0, P_{\text{BERT}_\theta}(y | c)\right) \right]\,,$$
where $\mathcal{L}$ is the cross-entropy loss function and $\Tilde{y}$ is either the ground truth $y_\text{gt}$ or $y_o$ in \eqref{eq:y_o}. As we will show in Section \ref{sec:results}, using $\Tilde{y}=y_o$ as the target is more reliable than $\Tilde{y}=y_\text{gt}$. Indeed, $y_\text{gt}$ is sometimes very distinct from the predicted transcripts in $Y$, which makes the task rather easy and far from reality at inference time.

The sum in the second term is over the remaining transcripts in $Y$. To avoid a high imbalance, we sum over only two randomly sampled transcripts with strictly higher WER than $y_o$, which yields a better performance on test.

%%%%%%%%%%%%%%%%%%%%%%%%%%%%%%%%%%%%%%%%%%%%
%%%%%%%%%%%%%%%%%%%%%%%%%%%%%%%%%%%%%%%%%%%%

\section{Data}
\label{sec:data}

As mentioned earlier, our research on LMs focuses on the low-resource data domain with respect to three dimensions: language (Norwegian), tone (spontaneous, conversational), and topics (parliament proceedings and customer service phone calls). The NST dataset used to train the primary AM is however more standard in tone and topics, consisting of grammatically correct, planned, noise-free speech.

A particularly challenging aspect of the Norwegian language is that it has two official written standards, a notably rich spectrum of spoken dialects \citep{SkjekkelandMartin2010D}, and no unequivocal correspondence between many dialect words and the official written form. Thus, transcribers usually follow certain conventions to ensure consistency and produce transcriptions as faithful as possible to the speech while adapting to the written standards. This is one of the core problems in ASR for Norwegian, in addition to data scarcity. The datasets used are described below and relevant properties are displayed in \autoref{tab:datasets}:

\newpage
%\medskip
\noindent
{\it NPSC: Norwegian Parliamentary Speech Corpus.} This public\footnote{Datasets and documentation are available from \url{https://www.nb.no/sprakbanken/en/resource-catalogue/oai-nb-no-sbr-58/}. We note that at the time we performed this work, only the beta version was available, containing 58h in contrast to the 140h  available at the time of writing.} dataset contains 58h of official proceedings from plenary meetings at the Norwegian parliament, reduced to 52h after cleaning. Recordings and transcriptions are split into sentences. Importantly, inaudible and/or unintelligible sounds and words, as well as fillings and hesitations, are transcribed as the blank token\footnote{This is not to be confused with the blank token used by the CTC decoder, which signals a separation between tokens.}. The split in train, evaluation and test sets was made by placing entire plenary sessions into different sets so that the topics also are split as much as possible, making the test closer to real situations.

\medskip\noindent
{\it TNCS: Telenor Norway Customer Service.} This internal dataset contains 149 conversations with the Norwegian customer service call center of Telenor. Recordings are made in two channels (customer and agent). For each channel, we divide the audio into shorter utterances\footnote{We use the WebRTC VAD Python package available from \url{https://github.com/wiseman/py-webrtcvad}.}. There is a substantial amount of overlapping speech and interruptions, making the sentence order ambiguous, which is especially relevant when training BERT with conversational context. Furthermore, that and the spontaneous tone cause the data to have a highly ungrammatical structure, making the LMs' task much harder. In addition, names, addresses and telecommunications-specific terms on the test data are mostly OOV tokens, so in that case the decoder outputs out-of-context candidates due to vocabulary constraints.
Noise, mumbling or unintelligible speech is transcribed as the blank token. The train-eval-test split was made by placing entire conversations into different sets.

%%%%%%%%%%%%%
\begin{table}[t!]
  \centering 
    \begin{tabular}{cccc}
       & \bf NPSC & \bf \ TNCS \ & \bf NST \\
    \hline
    \begin{tabular}{>{\centering}p{0.25\textwidth}c@{}} Number of utterances\end{tabular}
    \begin{tabular}{@{}c@{}} {\footnotesize \ Train\ } \\
    {\footnotesize \ Eval\ } \\
    {\footnotesize \ Test\ } \end{tabular} & 
    \begin{tabular}{@{}c@{}} {\footnotesize 23720} \\
    {\footnotesize 1279} \\
    {\footnotesize 1378} \end{tabular} & 
    \begin{tabular}{@{}c@{}} {\footnotesize 8532} \\
    {\footnotesize 1065} \\
    {\footnotesize 940} \end{tabular} & 
    \begin{tabular}{@{}c@{}} {\footnotesize 180237} \\
    {\footnotesize 20321} \\
    {\footnotesize 3370} \end{tabular} \\
    \hline
    \begin{tabular}{>{\centering}p{0.25\textwidth}c@{}} Average tokens per utterance\end{tabular}
    \begin{tabular}{@{}c@{}} {\footnotesize \ Train\ } \\
    {\footnotesize \ Eval\ } \\
    {\footnotesize \ Test\ } \end{tabular} & 
    \begin{tabular}{@{}c@{}} {\footnotesize 18.77} \\
    {\footnotesize 22.93} \\
    {\footnotesize 18.9} \end{tabular} & 
    \begin{tabular}{@{}c@{}} {\footnotesize 11.46} \\
    {\footnotesize 10.96} \\
    {\footnotesize 10.66} \end{tabular} & 
    \begin{tabular}{@{}c@{}} {\footnotesize 11.54} \\
    {\footnotesize 10.05} \\
    {\footnotesize 11.70} \end{tabular} \\
    \hline\hline
    \footnotesize Blank tokens (\%) & \footnotesize 2.44 & \footnotesize 5.36 & \footnotesize 0 \\
    %\hline
    \footnotesize Vocabulary size $|{\cal V}|$ & \footnotesize 26890 & \footnotesize 4368 & \footnotesize 78221 \\
    %\hline
    \footnotesize OOV test (\%) & \footnotesize 2.77 & \footnotesize 5.24 & \footnotesize 0.95 \\
    \hline
    \end{tabular}
    \caption{Data specifications. A ``blank token" is used to transcribe unintelligible sounds, fillings, and hesitations. ``Vocabulary size" denotes the number of unique words in the data. ``OOV test" is the percentage of words in the test set not found in the rest of the dataset (out of vocabulary).}
  \label{tab:datasets}
\end{table}
%%%%%%%%%%%%%%

One important feature of the TNCS data is that the distribution of number of tokens per utterance is extremely skewed towards low values and with a very large tail, despite having an average similar to NST data (see \autoref{tab:datasets}). This is natural for customer service conversations where quick turn-taking occurs, as well as long explanations. We will see this has a notable impact on our results. 

\medskip\noindent
{\it NST: Nordic Speech Technology.} This public\footnote{Datasets and documentation available at \url{https://www.nb.no/sprakbanken/en/resource-catalogue/oai-nb-no-sbr-13/}.} dataset is only used to train the primary AM, hence we just describe the relevant details. After processing we retain 339h of data divided into (300, 33, 6) hours for (train, evaluation, test) sets. In particular, for train and evaluation sets, we remove utterances with less than three words and those just containing three repetitions of the same word. For testing we select only fully grammatical, complete sentences.

%%%%%%%%%%%%%%%%%%%%%%%%%%%%%%%%%%%%%%%%%%%%
%%%%%%%%%%%%%%%%%%%%%%%%%%%%%%%%%%%%%%%%%%%%

\section{Experiments and results}
\label{sec:results}

We test several variations of shallow fusion with n-gram language models and N-best rescoring strategies with BERT. First, we train a baseline AM with greedy decoding\footnote{The optimization scheme of the AM is based on constant learning rate annealing, hence does not depend on the decoding strategy.} by fine-tuning the primary AM described in Section \ref{sec:baseline}. Next, we replace the greedy decoding with beam search, which is run with and without n-gram fusion. Then, we rescore the N-best lists from each beam search output using several BERT models trained on different tasks. We publish our source code\footnote{\url{https://gitlab.com/sburud/master}.} to make our results reproducible and to encourage further research. 

Our results report the total WER as the total edit distance for all samples divided by the total number of words. This penalizes every mistake equally as opposed to averaged utterance-level WER, that places higher penalties on errors in short transcripts.
This distinction is particularly important for our results on the TNCS dataset, where a large portion of the transcripts are very short. For N-best rescoring, we use the WER recovery rate (WERR) to measure improvements with respect to the gold standard (oracle). In the context of N-best rescoring as described in Section \ref{sec:integration}, it is defined as

\be
\text{WERR}(y')= \frac{\text{WER}(y_*) - \text{WER}(y')}{\text{WER}(y_*) - \text{WER}(y_o)}\ , 
\label{eq:werr}
\ee
where $y_o$ is the closest transcript to the ground truth and $y_*, y'$ are the top-ranked transcripts before and after rescoring, respectively. WER($\cdot$) is measured with respect to the ground truth $y_\text{gt}$.

%%%%%%%%%%%%%%%%%%%%%%%%%%%%%
\subsection{Setup}

For the beam search, we use a highly optimized, data-parallel beam search implementation\footnote{\url{https://github.com/parlance/ctcdecode}.}. We represent the beam search as a prefix tree, denoting the N highest-ranking nodes ``beam nodes''. At each time step, all non-beam nodes without beam node descendants are pruned.

At every time step, we consider adding every possible character to every beam node, except for characters only leading to OOV words. If the new character is non-blank\footnote{In CTC decoding, a blank token is used to separate tokens, see \citep{graves_connectionist_2006} for details. Not to be confused with the blank tokens used in the transcriptions as described in Section \ref{sec:data}.}, we add the n-gram and word count bonuses.

For BERT, we use a batch size of 768 and otherwise apply the hyperparameters listed under $\text{BERT}_\text{base}$ in \citep{devlin_bert_2019}. Due to the difference in size and topics across our data, training on multiple datasets simultaneously negatively impacts performance. Hence, when we list a BERT model trained on multiple datasets, this is to be understood as sequential training.

%%%%%%%%%%%%%%%%%%%%%%%%%%%%%%%
\subsection{Hyperparameter tuning}

We perform hyperparameter tuning in two stages: first the beam search parameters, and then the rescoring interpolation parameter, see Section \ref{sec:integration}.

For the beam search parameters, we perform different random searches for each task on the evaluation sets using Optuna \citep{akiba_optuna_2019}:
\begin{itemize}
    \item Tune $\alpha$ and $\beta$, minimizing $\sum \text{WED}(y_*, y_\text{gt})$. The parameters found are used to report baseline WER for shallow fusion.
    \item Tune $\alpha$ and $\beta$, minimizing $\sum \text{WED}(y_o, y_\text{gt})$. These parameters are used in rescoring experiments.
\end{itemize}
Using the optimal beam search parameters, we rescore all candidates in $Y$ using BERT. We then perform a grid search for the interpolation parameter $\gamma \in [0,0.5]$ with step size $0.001$.

%%%%%%%%%%%%%%%%%%%%%%%%%%%%%%%%%%%%
\subsection{N-best rescoring}
\label{sec:results_rescoring}

\autoref{tab:rescoring-bert-variants} shows rescoring performance for the various BERT fine-tuning strategies. For the sake of efficiency, not all experiments are performed on both datasets, since our  tests indicated that conclusions hold across our datasets.

On the one hand, the conversational NSP (C-NSP) objective does not help in rescoring, which we detail in Section \ref{sec:res-bert-text}. On the other hand, the disambiguation task improves performance significantly. This shows that more realistic training of transformer-based LMs does matter.
In line with \citet{devlin_bert_2019}, we found that BERT learns a sufficient language representation to adapt to new writing styles and language use with only minimal fine-tuning.

\begin{table}[t!]
\centering
\begin{tabular}{l|cc|cc}
\textbf{}                  & \multicolumn{2}{c|}{\textbf{TNCS}} & \multicolumn{2}{c}{\textbf{NPSC}} \\
\textbf{Decoding strategy} & \footnotesize\textbf{WER}    & \footnotesize\textbf{WERR}    & \footnotesize\textbf{WER}    & \footnotesize\textbf{WERR}   \\ \hline
Greedy decode              & 48.86            & -                & 37.70            & -               \\
BS with vocabulary           & 38.44            & -                & 30.70            & -               \\ \hline
BS with 2-gram               & 33.27            & -                & 23.39            & -               \\
+ Conversational NSP       & 33.27            & 0              & -               & -               \\
+ Disambiguation (no context)    & 32.99            & 3.33              & -            & -            \\
+ Disambiguation (short context) & \textbf{32.80}   & 5.60              & 22.54            & 11.99            \\
+ Disambiguation (long context)  & 33.09            & 2.14              & \textbf{20.75}   & 37.24            \\
+ \textit{Oracle rescorer} & \textit{24.87}   & \textit{100}   & \textit{16.30}   & \textit{100}  \\ \hline
BS with 6-gram               & 32.96            & -                & 23.05            & -               \\
+ Disambiguation (short context) & \textbf{32.74}   & 3.09              & 22.26            & 11.37            \\
+ Disambiguation (long context)  & -               & -                & \textbf{20.64}   & 34.68            \\
+ \textit{Oracle rescorer} & \textit{25.84}   & \textit{100}   & \textit{16.10}   & \textit{100} 
\end{tabular}
\caption{Results obtained with the different decoding strategies on test sets. Note that the value of $\gamma$ used to report results on the \emph{test} data is obtained from minimizing WER on the \emph{evaluation} data for each experiment. We use the total WER evaluated on the test split of each dataset, and report numbers as percentages. WERR is calculated for each block using the plain BS + n-gram model as baseline and the corresponding oracle rescorer as the gold standard, see \eqref{eq:werr}.}
\label{tab:rescoring-bert-variants}
\end{table}

The fact that the optimal value of $\gamma$ for test and evaluation is different is due to the small size of the datasets, which causes a significant variance among the splits. 

%%%%%%%%%%%%%%%%%%%%%%%%%%%%%%%%%%%%%%%%%%
\subsubsection{Disambiguation with conversational context}
\label{sec:res-context}
Much of the rescoring improvements originate from the presence of conversational context. As shown in \autoref{tab:rescoring-bert-variants}, introducing just two utterances of conversational context already improves rescoring results.

\begin{figure}[h!]
    \centering
    %\begin{subfig}
    %\centering
    \includegraphics[trim={0.1cm 0 1.1cm 0.5cm},clip,width=.63\linewidth]{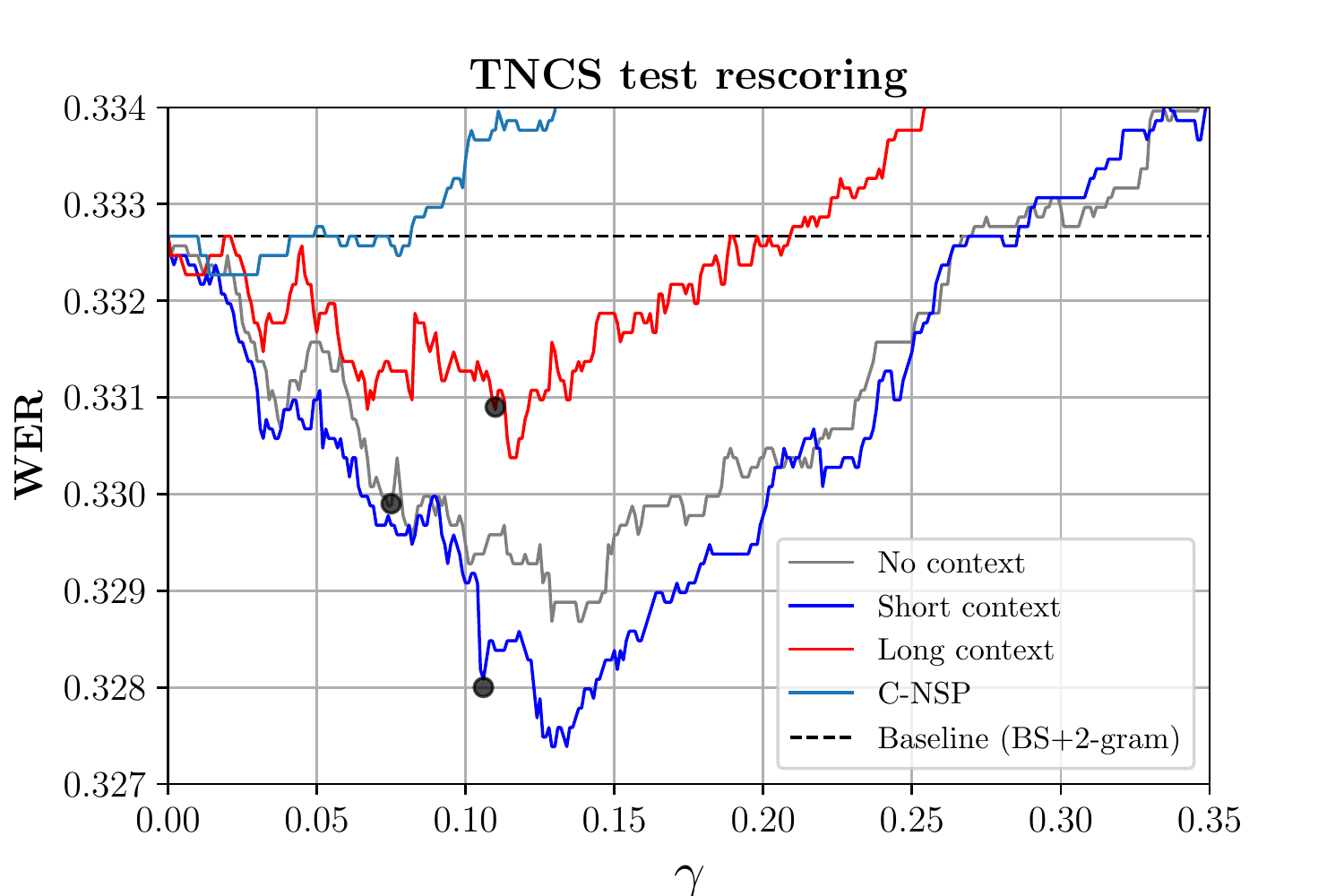}
    \caption{WER as a function of $\gamma$ on the TNCS test set. Reported values (grey dots) correspond to those in \autoref{tab:rescoring-bert-variants}, where the value of $\gamma$ is given by optimizing on the evaluation set. In the case of C-NSP, no improvement is reported since performance is consistently worse than the baseline on the evaluation set, even when small gains are observed on test.}
    \label{fig:tnn-extweight-test}
    %\end{subfig}%
\end{figure}

%trim={left bottom right top}
\begin{figure}[h!]
    \centering
    %\begin{subfig}
    %\centering
    \includegraphics[trim={0.1cm 0 1.1cm 0.5cm},clip,width=.63\linewidth]{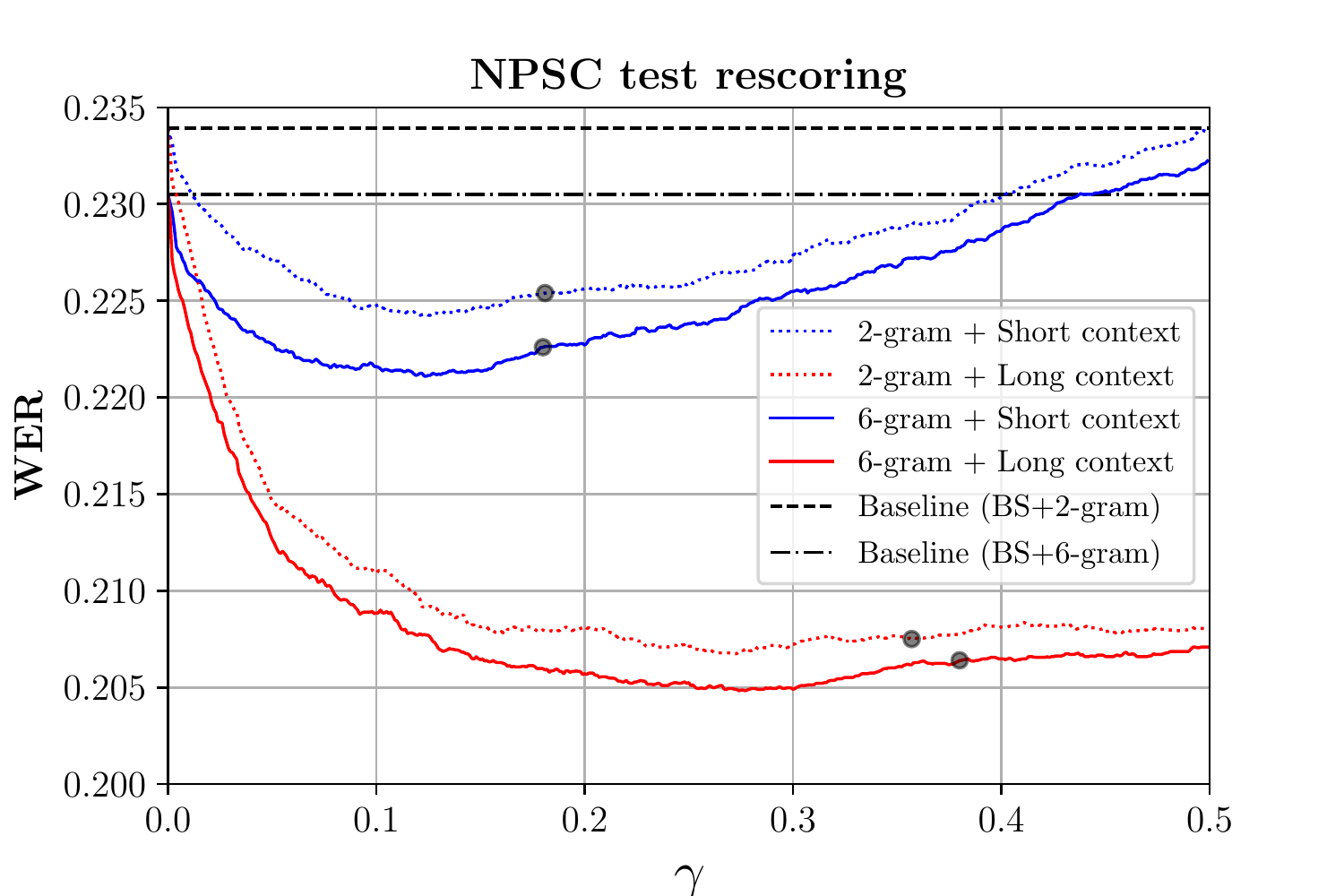}
    \caption{WER as a function of $\gamma$ on the NPSC test set. Grey dots correspond to reported values in \autoref{tab:rescoring-bert-variants}, given by the optima on evaluation data.}
    \label{fig:npsc-extweight-test}
    %\end{subfig}
    %\caption{Total caption}
\end{figure}

For the TNCS data, extending the context to five utterances reduced performance with respect to short (or no) context, see \autoref{fig:tnn-extweight-test}. Indeed, for unstructured conversations with quick turn-taking and topic changes, too long a context misleads BERT.

In contrast, NPSC benefits greatly from the longer context, obtaining 37.24\% and 34.68\% WERR after rescoring 2- and 6-gram shallow fusion models, respectively. With the longer context, significantly higher values for the interpolation weight $\gamma$ are favoured, as shown in \autoref{fig:npsc-extweight-test}. The improvements plateaued around $\gamma \approx 0.2$, but unlike the other experiments, a much higher $\gamma$ value was needed before we observed performance degradation. We attribute this to the long-form speeches being much more similar to the pre-training data. Further, the debates focus on a specific topic and are more structured, making it more likely that previous context is relevant when rescoring the current hypotheses.

\subsubsection{Sequence length and bounds for improvements}

\begin{figure}[h!]
\centering
    \includegraphics[trim={0.1cm 0 1.1cm 0.5cm},clip,width=.63\linewidth]{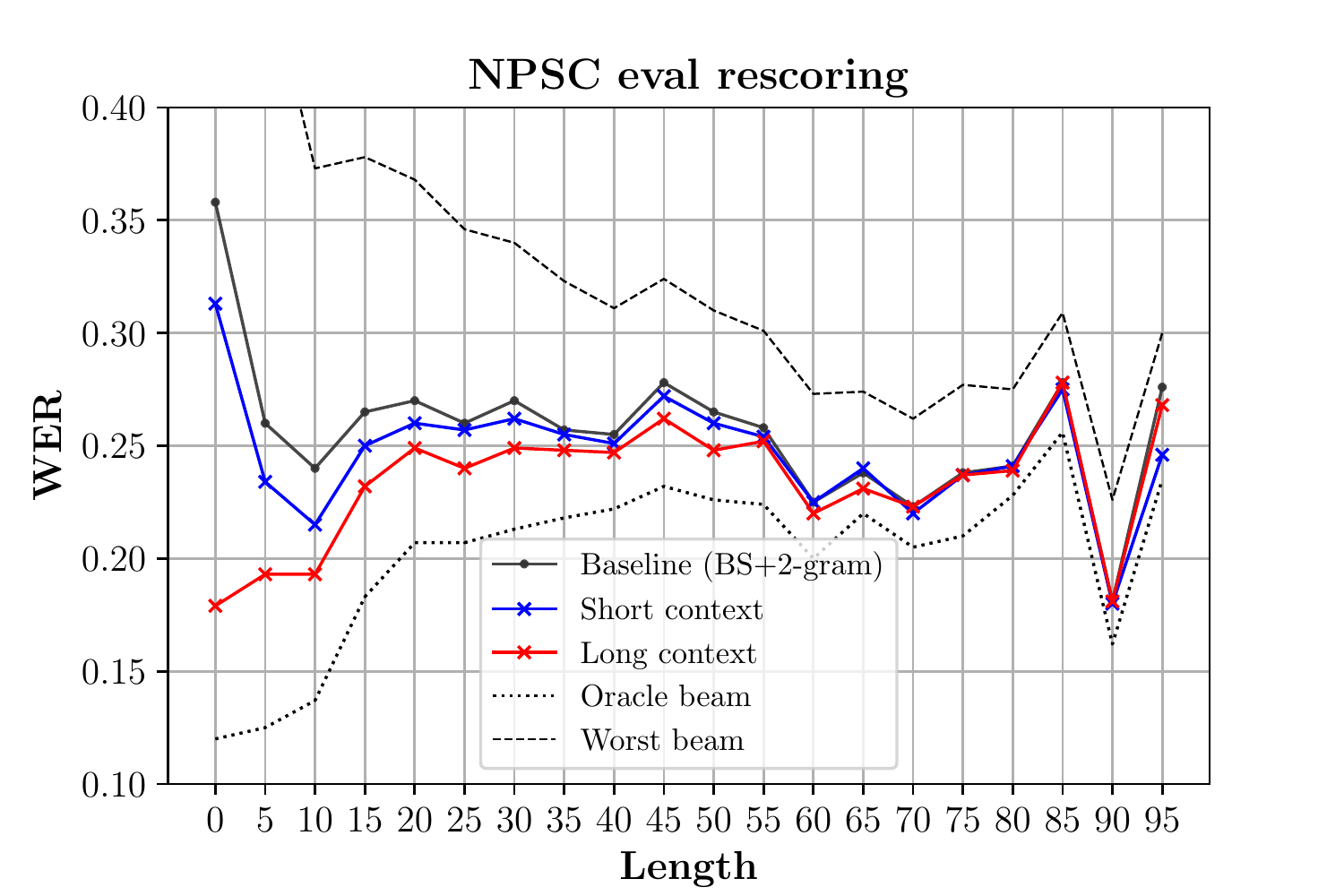}
    \caption{Total WER as a function of $y_\text{gt}$ word count (length) for the NPSC evaluation set. Bin $i$ denotes lengths within the interval $[i,i+4]$. ``Short (long) context'' rescores the baseline candidates with a disambiguation-BERT model fine-tuned with two (five) context utterances.}
    \label{fig:wer_by_length}
\end{figure}

For long utterances, the ratio between potential transcripts and N grows quickly, making the gap between upper and lower WER bounds rather narrow. On the other hand, CTC beam search will perform a near exhaustive search for short utterances, giving a much wider gap. This can be seen in \autoref{fig:wer_by_length}, which shows the empirical WER bounds as a function of utterance length, along with WER$(y_*)$ (baseline), and WER$(y')$, obtained after N-best rescoring with BERT fine-tuned on disambiguation with short or long context.

When increasing the context size available to BERT from two to five utterances, \autoref{fig:wer_by_length} clearly shows that the results improve more for shorter than longer utterances. Our interpretation is that short utterances are less prone to topic changes, often containing ans\-wers or follow-up questions, and thus are more consistent with the context. If we add to this having large part of the search space available, the task for BERT becomes easier.

For long sequences the gap between the best and worst transcripts is rather narrow due to the little variation among candidates. This severely limits BERT's ability to improve the results.

%%%%%%%%%%%%%%%%%%%%%%%%%%%%%%%%%%%%%%%%%%%
\bigskip
\subsection{ASR-independent disambiguation with BERT}
\label{sec:res-bert-text}

We have explored different training schemes for BERT. In Section \ref{sec:results_rescoring} we presented our results from integrating BERT as an LM for rescoring, and evaluated the WERR on ASR. Here we provide a deeper analysis of how those BERT fine-tuning strategies perform during disambiguation of ASR hypotheses \emph{independently} of the scores given by the AM and n-gram LM.

In order to do that, we take the BERT models fine-tuned on C-NSP (see Section \ref{sec:cnsp}) or disambiguation (see Section \ref{sec:disambiguation}) with short context and give them two ASR hypotheses: $y_o$ and another one from within the same beam. Then we measure their accuracy on the task of identifying $y_o$ under different training conditions. The difference with respect to N-best rescoring is that we only provide two candidates and a flat prior (same initial score).

Our findings are summarized in \autoref{tab:nsp-results}. Naturally, the (C-)NSP objective fails in the disambiguation task: the context representations learnt during (C-)NSP training are not useful for disambiguation, as the task at hand is substantially different. This is the main motivation to introduce the disambiguation task for rescoring. However we will see below that C-NSP almost reaches human accuracy at the task it is trained for, even when fine-tuned with small and noisy data such as our TNCS set.

\begin{table}[t!]
\centering
\begin{tabular}{lccc}
\textbf{Training data/objective\ } & \textbf{\ Acc\ } & \textbf{\ TPR\ } & \textbf{\ TNR\ } \\
Base model NSP     & 47.95           & 92.27& 3.63    \\
TNCS C-NSP          & 52.33           & 52.98      & 51.68      \\
TNCS Disambiguation ($y_o$)           & 80.96           & 91.71      & 70.20      \\
NPSC Disambiguation ($y_o$)           & 74.86           & 87.24      & 62.48      \\
+ TNCS Disambiguation ($y_o$)   & \textbf{82.17}  & 89.94      & 74.39      \\
+ TNCS Disambiguation ($y_\text{gt}$) & 79.05           & 81.25      & 75.00     
\end{tabular}
\caption{Text-only disambiguation accuracy on a balanced TNCS evaluation set with 2-utterance context. TPR/TNR are the true positive and negative rate, respectively. All values are given in percentage points.}
\label{tab:nsp-results}
\end{table}

On the other hand, disambiguation fine-tuning leads to substantial improvements. Results are, as expected, better when the training data is from the same dataset as the eva\-lu\-a\-tion set (TNCS), but still we see a significant uplift when training only on NPSC with respect to the base model. Further training on conversational data from TNCS consistently improves results. As argued earlier, using the oracle transcript $y_o$ during training is better than using the ground truth $y_\text{gt}$, since the former situation is more similar to the task at hand.

These analyses strengthen the rescoring results presented in Section \ref{sec:res-context} and shed light on their interpretation. Despite the high accuracy of disambiguation-BERT on this task, during N-best rescoring the amount of negative samples is much larger. More importantly, BERT's score is little significant when the AM and n-gram LM provide confident predictions, even if they are wrong. This results in a relatively low optimal value for $\gamma$ especially on the TNCS data. 
Conversely, higher $\gamma$ values lead to BERT overriding the AM and n-gram predictions in favour of likely incorrect predictions based purely on language and conversational context, ignoring what is actually said.

%%%%%%%%%%%%%%%%%%%%%%%%%%%%%%%%%%%%%%%%%%%%%%%%%%%%%
\subsubsection{Conversational NSP and human performance}

We now evaluate C-NSP fine-tuned BERT models on the task they were trained for, and analyze how the nature of the training data affects their performance on TNCS data. As shown in \autoref{tab:conv-nsp}, the accuracy notably increases once BERT is exposed to the relevant conversational data, even with such a little amount of it.

\begin{table}[b!]
\centering
\begin{tabular}{lc}
\textbf{C-NSP model\ \ \ }    & \textbf{\ Acc\ } \\
\hspace{1em} Base           & 53.13         \\
\hspace{1em} NPSC           & 59.38         \\
\hspace{1em} TNCS           & 68.75         \\
\hspace{1em} Human          & 70.31         
\end{tabular}
\caption{C-NSP accuracy (\%) on 64 evaluation samples from a balanced C-NSP set from TNCS evaluation data. The human was not fine-tuned on this task nor exposed to the evaluation samples beforehand.}
\label{tab:conv-nsp}
\end{table}

When we attempted the C-NSP task ourselves, we found it to be surprisingly tricky. Most samples required heavy use of domain knowledge, both in the form of common patterns in phone conversations and facts specific to telecommunications. This makes the results of C-NSP quite impressive, nearly reaching human accuracy on such highly unstructured conversational data, although the learnt representations help little in transcript disambiguation and rescoring.

%%%%%%%%%%%%%%%%%%%%%%%%%%%%%%%%%%%%%%%%%%%%%%%%%%%%%
%%%%%%%%%%%%%%%%%%%%%%%%%%%%%%%%%%%%%%%%%%%%%%%%%%%%%

\section{Conclusions and future work}
\label{sec:conclusions}

We presented novel, efficient techniques to rescore ASR candidates using a large pre-trained BERT model.
These techniques require little fine-tuning and are particularly useful in the low-resource domain, since we only fine-tune with the few training transcriptions available for ASR.

In particular, disambiguation-BERT makes use of past conversational context and is trained to identify the best transcript among hypotheses. Applying it for rescoring leverages WER relative recoveries of over 37\%
when transcribing parliamentary debates. For spontaneous phone conversations improvements are more modest due to the quick turn-taking and topic changes, which also makes the model prefer less conversational context.

In the case of conversational NSP, the task is rather difficult for humans as well. Even so, a BERT model fine-tuned on C-NSP is only 2.2\% worse than a human, although the language representations learnt are not as useful for the rescoring of ASR hypotheses. 

Our analyses also reveal that the contribution of disambiguation-BERT in rescoring could be further optimized, since $\gamma$ is optimized given a whole evaluation set rather than for each utterance. That is, one could identify when the beam search scores from the AM and n-gram LM have low confidence, and set a higher weight on BERT's contribution in those cases. Such tighter integration will be the subject of future work.

For long utterances, ASR hypotheses are very similar to each other, since early variations are more likely to be pruned during decoding. This makes the task or the LMs much harder and limits their potential. We did some preliminary investigations\footnote{\url{https://gitlab.com/sburud/master/-/tree/master/speechlm/ctc-decode}.} on the inclusion of a diversity bonus in the beam search inspired by the work of \citet{vijayakumar_diverse_2018} in image captioning, with the aim of forcing the beam search to explore different modes and increase the variability among candidates. This could potentially give higher-quality beams and enable, among other benefits, rescoring improvements independently of the rescoring technique used. In future work we will report on our investigations and the effect of beam diversity in ASR.

In this work we have exploited the large gap between the top-ranked transcription and the oracle transcript in low-resource conversational ASR. Our findings contribute to reducing that gap without modifying the acoustic model, although modifications of that component will also be investigated in future work. Of particular interest in our low-resource domain are end-to-end architectures that make better use of the information contained in the (limited) data available for ASR training, in addition to exploiting the language representations provided by large pre-trained language models.

%%%%%%%%%%%%%%%%%%%%%%%%%%%%%%%%%%%%%%%%%%%%
%%%%%%%%%%%%%%%%%%%%%%%%%%%%%%%%%%%%%%%%%%%%
\newpage
\acks{We are thankful for feedback and enlightening discussions with Knut Kvale, Ole Jakob Mengshoel, Massimiliano Ruocco, Giampiero Salvi, Marco Siniscalchi and Torbjørn Svendsen. 
We are very grateful to Knut Kvale for his valuable collaboration on transcribing Telenor conversational data. We also wish to thank Petr Taborsky for collaboration during the initial stages of this work on training the acoustic models, and to Weiqing Zhang for technical support related to the GPU servers.

This work was partially carried out as part of SB's master thesis \citep{simen_master}. PO has been partially supported by the Norwegian Research Council through the IKTPLUSS grant for the SCRIBE project\footnote{\url{https://scribe-project.github.io/}} (KSP21PD).}

%%%%%%%%%%%%%%%%%%%%%%%%%%%%%%%%%%%%%%%%%%%
%%%%%%%%%%%%%%%%%%%%%%%%%%%%%%%%%%%%%%%%%%%

\vskip 0.2in
\bibliography{main3.bib}

\end{document}